%% file: main.tex
\documentclass{article}
\usepackage{multirow, tabularx}
\usepackage{url, subfigure}
\usepackage{graphicx,times, paralist}
\usepackage{makecell, bm, amsbsy}
\usepackage{color, graphicx}
\usepackage{float}
\usepackage{algorithmic,algorithm}
\usepackage[colorlinks,
linkcolor=red,
anchorcolor=blue,
citecolor=blue,
urlcolor=blue
]{hyperref}
\input{math_commands.tex}

    \PassOptionsToPackage{numbers, compress}{natbib}

\usepackage[preprint]{neurips_2018}

\usepackage[utf8]{inputenc} 
\usepackage[T1]{fontenc}    
\usepackage{hyperref}       
\usepackage{url}            
\usepackage{booktabs}       
\usepackage{amsfonts,amsmath}       
\usepackage{nicefrac}       
\usepackage{microtype}      
\usepackage{xcolor}

\title{Pre-Training Graph Neural Networks for\\ Generic Structural Feature Extraction}

\author{Ziniu Hu, Changjun Fan, Ting Chen, Kai-Wei Chang, Yizhou Sun\\
  University of California, Los Angeles\\
  \texttt{\{bull, cjfan2017, tingchen, kwchang, yzsun\}@cs.ucla.edu}
}

\begin{document}
\maketitle

\begin{abstract}
\input{section/abstract}
\end{abstract}
\section{Introduction}\label{sec:introduction}\input{section/introduction}
\section{Related Work}\label{sec:related}\input{section/related}
\section{Approach}\label{sec:approach}\input{section/approach}
\section{Experiments}\label{sec:evaluation}\input{section/evaluation}
\section{Conclusion}\label{sec:conclusion}\input{section/conclusion}

\bibliographystyle{plain}
\bibliography{main}

\newpage
\appendix
\input{section/appendix.tex}

\end{document}

%% file: math_commands.tex
\usepackage{amsmath,amsfonts,bm}

\def\eqref#1{equation~\ref{#1}}

\def\1{\bm{1}}

\newcommand{\Ab}{\mathbf{A}}

\newcommand{\Db}{\mathbf{D}}
\newcommand{\Eb}{\mathbf{E}}

\newcommand{\Hb}{\mathbf{H}}
\newcommand{\Ib}{\mathbf{I}}

\newcommand{\Rb}{\mathbf{R}}

\newcommand{\Wb}{\mathbf{W}}

\newcommand{\cA}{\mathcal{A}}

\newcommand{\cC}{\mathcal{C}}
\newcommand{\cD}{\mathcal{D}}
\newcommand{\cE}{\mathcal{E}}
\newcommand{\cF}{\mathcal{F}}
\newcommand{\cG}{\mathcal{G}}

\newcommand{\cL}{\mathcal{L}}

\newcommand{\cO}{\mathcal{O}}

\newcommand{\cU}{\mathcal{U}}
\newcommand{\cV}{\mathcal{V}}

\DeclareMathAlphabet{\mathsfit}{\encodingdefault}{\sfdefault}{m}{sl}
\SetMathAlphabet{\mathsfit}{bold}{\encodingdefault}{\sfdefault}{bx}{n}



%% file: section/abstract.tex
Graph neural networks (GNNs) are shown to be successful in modeling applications with graph structures. 
However, training an accurate GNN model requires a large collection of labeled data and expressive features, which might be inaccessible for some applications. 
To tackle this problem, we propose a pre-training framework that captures generic graph structural information that is transferable across tasks. Our framework can leverage the following three tasks:
1) denoising link reconstruction, 2) centrality score ranking, and 3) cluster preserving. The pre-training procedure can be conducted purely on the synthetic graphs, and the pre-trained GNN is then adapted for downstream applications. 
With the proposed pre-training procedure, the generic structural information is learned and preserved, thus the pre-trained GNN requires less amount of labeled data and fewer domain-specific features to achieve high performance on different downstream tasks.    
Comprehensive experiments demonstrate that our proposed framework can significantly enhance the performance of various tasks at the level of node, link, and graph.

%% file: section/introduction.tex
Graphs are a fundamental abstraction for modeling relational data in physics, biology, neuroscience and social science. 
Although there are numerous types of graph structures, some graphs are known to exhibit rich and generic connectivity patterns
that appear general in graphs associated with different applications. 
Taking network motifs, which are some small sub-graph structures, as an example, 
they are considered as the building blocks for many graph-related tasks~\cite{benson2016higher}, e.g., triangular motifs are crucial for social networks, two-hop paths are essential to understand air traffic patterns, etc.
Despite its importance, previous researchers are required to design various specific rules or patterns to extract motif structures, in order to serve as features for different applications manually. This process is tedious and ad-hoc.

Recently, researchers have adopted deep representation learning into graph domain and proposed various Graph Neural Network architectures~\cite{gcn, graphsage, DBLP:conf/iclr/VelickovicCCRLB18} to alleviate this issues by automatically capturing complex information of graph structures from data. 
In general, GNNs take a graph with attributes as input 
and apply convolution filters to generate node embeddings with different granularity levels layer by layer. 
The GNN framework is often trained in an end-to-end manner towards some specific tasks and has shown to achieve competitive performance in various graph-based applications, such as semi-supervised node classification~\cite{gcn}, recommendation systems~\cite{DBLP:conf/kdd/YingHCEHL18} and knowledge graphs~\cite{DBLP:conf/esws/SchlichtkrullKB18}.  

Despite the success, most of the GNN applications heavily rely on the domain-specific input features. For example, in PPI dataset~\cite{zitnik2017predicting}, which is widely used as a node classification benchmark task, positional gene sets, motif gene sets, and immunological signatures are used as features. 
However, these domain-specific features can be hard to obtain, and they cannot generalize to other tasks.
Without access to these domain-specific features, the performance of GNNs is often suboptimal.
Besides, unlike in the fields of computer vision (CV) or natural language processing (NLP), where large-scale labeled data are available \citep{deng2009imagenet}, in many graph-related applications, graphs with high-quality labels 
are expensive or even inaccessible. For example, researchers in neural science use fMRI to scan and construct brain networks~\cite{brain}, which is extremely time-consuming and costly. Consequently, the insufficiency of labeled graphs restricts the potential to train deep GNNs that learn generic graph features from scratch. To tackle these issues, in this paper, 
we consider training a deep GNN from a large set of unlabelled graphs and transferring the learned knowledge to downstream tasks with only a few labels, an idea known as pre-training. We focus on the following research questions: \emph{Can GNNs learn the generic graph structure information via pre-training? And to what extent can the pre-training benefit the downstream tasks?} 

Although the idea of unsupervised pre-training has been proved successful in CV and NLP~\cite{girshick2014rich, DBLP:journals/corr/abs-1810-04805}, adopting this idea for training GNNs is still a non-trivial problem. First, even for unlabeled 
graphs, collecting a high-quality and generalizable graph dataset for pre-training is still hard. Graphs in different domains usually possess different properties~\cite{snapnets}. Taking the degree distribution~\cite{barabasi2004network} as an example, it tends to be uniform in chemical molecular networks, while it always follows the power-law in social networks. Therefore, the model trained on graphs from one domain often cannot generalize well to another domain. As a consequence, obtaining a high-quality graph dataset that covers various graph properties is crucial and challenging for pre-training GNNs. To solve this, we propose to generate \emph{synthetic graphs} with different statistical properties as the pre-training data. Specifically, we exploit the degree-corrected stochastic block model~\cite{holland1983stochastic} to generate synthetic graphs, 
 where various of parameters are used in order to generate a variety of graphs with different properties.

Second, as is shown in~\cite{newman2010networks}, graphs have a wide spectrum of structural properties, ranging from nodes, edges to sub-graphs. Existing unsupervised objectives such as forcing the node embeddings to preserve the similarity derived from random walks~\cite{perozzi2014deepwalk, grover2016node2vec} only focus on capturing the relatively local structures and overlook the higher-level structural information. To capture information present in different levels of graph structures, we design three self-supervised pre-training tasks: 
1) denoising link reconstruction, 2) centrality score ranking, and 3) cluster preserving.
Guided by these tasks, the pre-trained GNNs are able to capture general graph properties and benefit the downstream tasks. 

Our main contributions are as follows:

\begin{compactitem}
\item We propose a pre-training framework that enables GNNs to learn generic graph structural features. The framework utilizes synthetic graphs with adjustable statistical properties as the pre-training data. Therefore, it does not require additional label data. The pre-trained model can benefit applications with unseen graphs.
\item We explore three self-supervised pre-training tasks which focus on different levels of graph structures, enabling GNNs to capture multi-perspective graph structure information.
\item We perform extensive experiments on different downstream tasks and demonstrate that all of them can be significantly benefited from transferring the knowledge learned via pre-training. We will release source code and data to facilitate future research on this line. 
\end{compactitem}

%% file: section/related.tex
The goal of pre-training is to allow a neural network model to initialize its parameters with weights learned from the pre-training tasks. In this way, the model can leverage the commonality between the pre-training and the downstream tasks. Pre-training has shown superior in boosting the performance of many downstream applications in computer vision (CV), natural language processing (NLP) and graph mining. In the following, we review approaches and applications of pre-training in these fields.
\paragraph{Pre-training strategies for graph applications} Previous studies have proposed to utilize pre-training to learn graph representations. These attempts  directly parameterize the node embedding vectors and optimize them by preserving some deterministic measures, such as the network proximity~\cite{tang2015line} or statistics derived from random walks~\cite{grover2016node2vec}. However, the embedding learned in this way cannot generalize to another unseen graph as the information they capture are graph-specified. In contrast, we consider a transfer learning setting, where our goal is to learn generic graph representations. 

With the increasing focus on graph neural networks (GNNs), researchers have explored the direction of pre-training GNNs on unannotated data. GraphSAGE~\cite{graphsage} adds an unsupervised loss by using random walk based similarity metric. Graph Infomax~\cite{DBLP:journals/corr/abs-1809-10341} 
proposes to maximize mutual information between node representations obtained from GNNs and a graph summary representation. All these studies only pre-train GNNs on a particular set of graphs with task-specific feature inputs. As the feature types across graph datasets are quite different, it is hard to generalize the knowledge learned from one set to another. Instead, our approach 
learns generic structural information of graphs, regardless of their specific feature types. 

\paragraph{Pre-training strategies for other machine learning applications} Pre-training in CV~\cite{girshick2014rich, zeiler2014visualizing,donahue2014decaf}  mostly follows the following paradigm: first pre-train a model on large-scale supervised datasets (such as ImageNet~\cite{deng2009imagenet}), then fine-tune the pre-trained model on downstream tasks~\cite{girshick2014rich} or directly extract the representation as features~\cite{donahue2014decaf}. 
This approach has been shown to make significant impact on various downstream tasks including object detection~\cite{girshick2014rich, he2017mask}, image segmentation~\cite{chen2018encoder, long2015fully}, etc.

Pre-training has also been used in various NLP tasks.
On the word level, word embedding models~\cite{mikolov2013distributed,pennington2014glove, bojanowski2017enriching} capture semantics of words by leveraging occurrence statistics on text corpus and have been used as a fundamental component in many NLP systems. On the sentence level, pre-training approaches have been applied to derive sentence representations~\cite{kiros2015skip, le2014distributed}.  Recently, contextualized word embeddings~\cite{DBLP:conf/naacl/PetersNIGCLZ18,DBLP:journals/corr/abs-1810-04805}
are proposed to pre-train a text encoder on large corpus with a language model objective to better encode words and their context. The approach has shown to reach state-of-the-art performance in multiple language understanding tasks on the GLUE benchmark~\cite{DBLP:conf/emnlp/WangSMHLB18} and on a question answering dataset~\cite{DBLP:conf/emnlp/RajpurkarZLL16}.

%% file: section/approach.tex
In this section, we first provide an overview of the proposed pre-training framework. Then, we introduce three self-supervised tasks, in which different types of structural graph information can be captured by the designed GNN. Finally, we discuss how to pre-train the GNN model on synthetic graphs and how to adapt the pre-trained GNN to downstream tasks.

\begin{figure*}[t]
\centering
\includegraphics[width=0.85\linewidth]{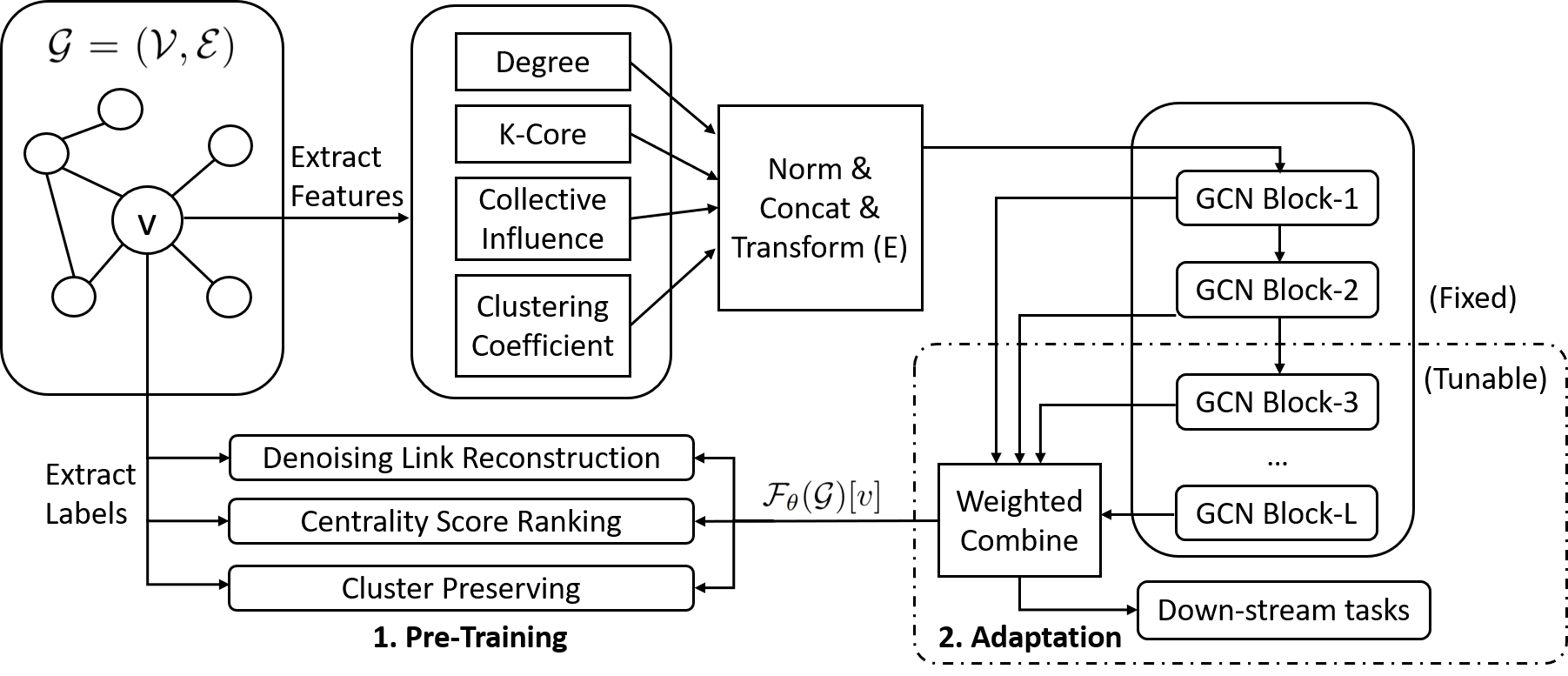}
\caption{The overall framework for pre-training and adaption. First, a multi-layer GNN is trained on three structure-guided tasks. Then,
the lower layer of the GNN is fixed and the upper layer of the GNN is fine-tuned on the given downstream task.}\label{fig:encoder}
\end{figure*}

\subsection{Pre-training Framework}
Since the goal of pre-training is to learn a good feature extractor and parameter initialization 
from pre-training tasks for downstream tasks, the model architecture and input attributes have to be consistent through these two phases. Based on this requirement, we design a pre-training framework, as is summarized in Figure~\ref{fig:encoder}. 

We consider the following encoder-decoder framework. Given a graph $\cG = (\cV,\cE)$, where $\cV$ and $\cE$ denote the set of nodes and edges. $\Ab$ denotes the adjacency matrix, which is normally sparse.
We encode each node in the graph by a multi-layer GNN encoder $\cF$ into a set of representation vectors $\cF(\cG)$. These node representations are then fed into task-specific decoders to perform the downstream tasks. 
If a large collection of labeled data is provided, the encoder $\cF$ and the task-specific decoder can be trained jointly in an end-to-end manner. However, the number of labeled data of some downstream tasks is often scarce in practice. Consequently, the performance by end-to-end training may be sub-optimal. In this paper, we propose to first pre-train the encoder $\cF$ on self-supervised tasks, which are designed for facilitating the encoder to capture generic graph structures. Note that, the node representation of an $L$-layer GNN's output utilizes the information from $L$-hop neighborhoods context. 
Thus, if pre-trained well, the encoder $\cF_{\theta^*}$ with learned weight $\theta^*$ can capture high-level graph structure information that cannot be easily represented by local features. We then cascade the resulted $\cF_{\theta^*}$ with downstream decoder and fine-tuned the model. 
We discuss the input representation and model architecture as follow:
\paragraph{Input Representation} 
As our goal is to enable GNNs to learn generalizable graph structural features, the input to the model should be generic to all graphs. Thus, we select four node-level features which can be computed efficiently. They are (1) {\it Degree}, which defines the local importance of a node; (2) {\it Core Number}, which defines the connectivity of a node's subgroup; (3) {\it Collective Influence}, which defines a node's neighborhood's importance; and (4) {\it Local Clustering Coefficient}, which defines the connectity of a node's 1-hop neighborhood\footnote{Detail descriptions and time complexity of these features are in Appendix}. 

To generalize these features to graphs with different sizes, we conduct a min-max normalization to them except the Local Clustering Coefficient (because it is already normalized). 
We then concatenate these four features and cascade them with a non-linear transformation $\Eb$ 
to encode the local features into a $d$-dimensional embedding vector, where $d$ is the input dimension to GNNs.

\paragraph{GNN Model Architecture}
As discussed above, our GNN architecture design should be (1) powerful enough to capture graph structural features at both local- and global-level; (2) general enough so that it can be pre-trained and transferred. Based on these conditions, we design a modified Graph Convolutional Networks (GCNs) operator as the basic building block and stack $L$ blocks together to construct the proposed architecture (denoted as GNN through this paper).
Similar to GCNs, our architecture considers a fixed convolutional filter as it makes the model easier to be transferred. 
Mathematically, our modified GCN block is defined as
\begin{align}
\Hb^{(l+1)} = \sigma\left( (\tilde{\Db}^{-\frac{1}{2}}\tilde{\Ab}\tilde{\Db}^{-\frac{1}{2}}) \left( {norm}\left(\sigma \left(\Hb^{(l)}\Wb_1^{(l)}\right)\Wb_2^{(l)}\right)\right)\right),
\end{align}
where $\Hb^{(l)}$ is the input of the $l$-th hidden layer of GNN, i.e., the output of the $(l-1)$-th hidden layer, $\tilde{\Db}^{-\frac{1}{2}}\tilde{\Ab}\tilde{\Db}^{-\frac{1}{2}}$ is the normalized Laplacian matrix of graph $\cG$ as the convolutional filter. $\tilde{\Ab}=\Ab+\Ib$ is the adjacency matrix with self-loop, which serves as skip-connection, and $\tilde{\Db}$ is a diagonal matrix, where 
$\tilde{\Db}_{i,i} = \sum_j \tilde{\Ab}_{i,j}$, representing degree of each node.
$\Wb_1^{(l)}$ and $\Wb_2^{(l)}$ are the weight matrix in the $l$-th hidden layer that would be trained, and $\sigma(\cdot)$ is the activation function. We choose two weight matrix as~\cite{DBLP:journals/corr/abs-1810-00826} points out that multi-layer perceptron can help GNNs learn better representation. 
$norm$ is the batch normalization~\cite{DBLP:conf/icml/IoffeS15} operation to help convergence.
In summary, for any given node, a GNN layer will transform each node by a two-layer perceptron $\Wb_1^{(l)}$, $\Wb_2^{(l)}$ with normalization $norm$, followed by an aggregation of the transformed embeddings of its neighbors with weights specified in $\tilde{\Db}^{-\frac{1}{2}}\tilde{\Ab}\tilde{\Db}^{-\frac{1}{2}}$.

To conduct different tasks using the output of GNNs, we need to extract a representation vector $\cF^{task}$ as the task's input. 
As prior studies~\cite{DBLP:conf/naacl/PetersNIGCLZ18} observe that different tasks require information from different layers, we consider $\cF^{task}$ as a linear combination of node representations from different GNNs layers $\{\Hb^{(l)}\}_{l=1}^{L}$, 
so that different tasks can utilize different perspective of structural information:
\begin{align}
\cF^{task} = \alpha^{task} \sum_{l=1}^{L} \beta^{task}_l \cdot \Hb^{l}, \ \ \text{where} \ \ \beta^{task}_l = \frac{\exp(\psi^{task}_l)}{\sum_{i=1}^{L} \exp(\psi^{task}_i)},
\end{align}
where $\alpha^{task}$ is a scaling vector for ease of optimization, and $\beta^{task}$ is softmax-normalized vector gotten from $\psi$ to give different attention to different layers. The obtained $\cF^{task}$ is then fed to different task-specific decoder to conduct different tasks. Different from~\cite{DBLP:conf/naacl/PetersNIGCLZ18}, which considers only a single pre-training task, we have multiple pre-training tasks that focus on different aspects of the graph structure. Therefore, we use the same weight combination architecture and maintain different sets of $\alpha^{task}$ and $\beta^{task}$ for different pre-training tasks, which will also be learned. 
\subsection{Self-supervised Pre-training Tasks}
As is shown in~\cite{newman2010networks}, graphs have a wide spectrum of structural properties, ranging from nodes, edges to sub-graphs. Based on this, we design three self-supervised tasks that focus on different perspectives of graph structure to pretrain the GNN model. In the following, we use $\cF^{task}(\cG)[v]$ to denote the representation for node $v$ in $\cG$ for Task indexed by $task$. Note for each task, $\cF^{task}(\cG)$ shares the same GCN parameter $\theta$, but has different scaling parameter $\alpha^{task}$ and mixing parameter $\beta^{task}$. 

\paragraph{Task 1: Denoising Link Reconstruction}
A good feature extractor should be able to restore links even if they are removed from the given graph. Based on this intuition, we proposed denoising link reconstruction (i.e., $rec$). 
Specifically, we add noise to an input graph $\cG$ and obtain its noised version $\cG^*$ by randomly removing a fraction of existing edges. The GNN model takes the noised graph $\cG^*$ as input and learns to represent the noisy graph as $\cF^{rec}(\cG^*)$. The learned representation is then passed to a neural tensor network (NTN)~\cite{DBLP:conf/nips/SocherCMN13} pairwise decoder $\cD^{rec}(\cdot, \cdot)$ 
, which predicts if two nodes $u$ and $v$ are connected or not as:
$\hat{\Ab}_{u,v} = \cD^{rec}\Big(\cF^{rec}(\cG^*)[u], \cF^{rec}(\cG^*)[v]\Big)$.
Both encoder $\cF^{rec}$ and decoder $\cD^{rec}$ are optimized jointly with an empirical binary cross-entropy loss:
$$
    \cL_{rec} = -\sum\nolimits_{u, v \in \cV} \Big(\Ab_{u,v} \log(\hat{\Ab}_{u,v}) + (1-\Ab_{u,v}) \log(1-\hat{\Ab}_{u,v}) \Big).
$$ 
In this way, the pre-trained GNNs is able to learn a robust representation of the input data~\cite{DBLP:journals/jmlr/VincentLLBM10}, which is especially useful for incomplete or noisy graphs. Since we want such capacity to deal with noisy graph maintained in the other two tasks, the following two tasks also take the noised graph $\cG^*$ as input.

\paragraph{Task 2: Centrality Score Ranking} 
Node centrality is an important metric for graphs~\cite{10.2307/2780000}. It measures the importance of nodes based on their structural roles in the whole graph. Based on this, we propose to pre-train GNNs to rank nodes by their centrality scores (i.e., $rank$), so the pre-trained GNNs can capture structural roles of each node. Specifically, four centrality scores are used: 
\begin{compactitem}
\item [i] {\it Eigencentrality}~\citep{eigen} measures nodes' influences based on the idea that high-score nodes contribute more to their neighbors. It describes a node's `hub' role in a whole graph.
\item [ii] {\it Betweenness}~\citep{freeman1977set} measures the number of times a node lies on the shortest path between other nodes. It describes a node's `bridge' role in a whole graph.
\item [iii] {\it Closeness}~\citep{closeness} measures the total length of the shortest paths between one node and the others. It describes a node's `broadcaster' role in a whole graph.
\item [iv] {\it Subgraph Centrality}~\citep{estrada2005subgraph} measures the the participation of each node (sum of closed walks) in all subgraphs in a network. It describes a node's `motif' role in a whole graph.
\end{compactitem}
The above four centrality scores concentrate on the different roles of a node in the whole graph. Thus, a GNN that can estimate these scores is able to preserve multi-perspective global information of the graph.
Since centrality scores are not comparable among different graphs with different scales, we resort to rank the relative orders between nodes in the same graph regarding the three centrality scores. For a node pair ($u, v$) and a centrality score $s$, with relative order as $\Rb_{u,v}^s=(s_u > s_v)$, a decoder $\cD^{rank}_s(\cdot)$ for centrality score $s$ estimates its rank score by $\hat{S}_{v} = \cD^{rank}_s(\cF^{rank}(\cG^*)[v])$. Following the pairwise ranking setting defined in~\cite{burges2005learning}, the probability of estimated rank order is defined by $\hat{\Rb}_{u,v}^s = \frac{\exp(\hat{S}_{u}-\hat{S}_{v})}{1+\exp(\hat{S}_{u}-\hat{S}_{v})}$. We optimize $\cF^{rank}$ and $\cD^{rank}_s$ for each centrality score $s$ by:
$$
    \cL_{rank} = -\sum\nolimits_{s}\sum\nolimits_{u, v \in \cV} \Big(\Rb_{u, v}^s \log(\hat{\Rb}_{u, v}^s) + (1-\Rb_{u, v}^s) \log(1-\hat{\Rb}_{u, v}^s) \Big).
$$
In this way, the pre-trained GNNs can learn the global roles of each node in the whole graph. %

\paragraph{Task 3: Cluster Preserving}
One important characteristics of real graphs is the cluster structure \citep{DBLP:journals/csr/Schaeffer07}, meaning that nodes in a graph have denser connections within clusters than inter clusters. We suppose the nodes in a graph is grouped into $K$ different non-overlapping clusters $\cC=\{C_i\}_{i=1}^{K}$, where $\cup_i C_i=\cV$ and $\forall C_i, C_j \in \cC: C_i \cap C_j = \emptyset$.
We also suppose there is an indicator function $\Ib(\cdot)$ to tell which cluster $C$ a given node v belongs to. Based on $\Ib$, we then pre-train GNNs to learn node representations so that the cluster information is preserved (i.e., $cluster$).
First, we use an attention-based aggregator $\cA$ to get a cluster representation by $\cA\big(\{\cF^{cluster}(\cG^*)[v]\ |\ v  \in C\}\big)$. Then, a neural tensor network (NTN)~\cite{DBLP:conf/nips/SocherCMN13}
decoder $\cD^{cluster}(\cdot, \cdot)$ estimates the similarity of node $v$ with cluster $C$ by: $S(v, C) = \cD^{cluster}\bigg(\cF^{cluster}(\cG^*)[v], \ \cA\Big(\{\cF^{cluster}(\cG^*)[v]\ |\ v  \in C\}\Big)\bigg)$. We then estimate the probability that $v$ belongs to $C$ by $P(v \in C) = \frac{\exp\big(S(v, C)\big)}{\sum_{C' \in \cC} \exp\big(S(v, C')\big)}$. Finally, we optimize $\cF^{cluster}$ and $\cD^{cluster}$ by:
    $\cL_{cluster} = -\sum\nolimits_{v \in \cV} \Ib(v) \log\bigg(P\big(v \in \Ib(v)\big)\bigg)$.

The pre-trained GNNs can learn to embed nodes in a graph to a representation space that can preserve the cluster information, which can be easily detected by $\cD^{cluster}$. 
\begin{figure*}[t!]
    \subfigure[Dense and Power-law]{
        \label{fig:d_s}
        \includegraphics[width=0.22\textwidth]{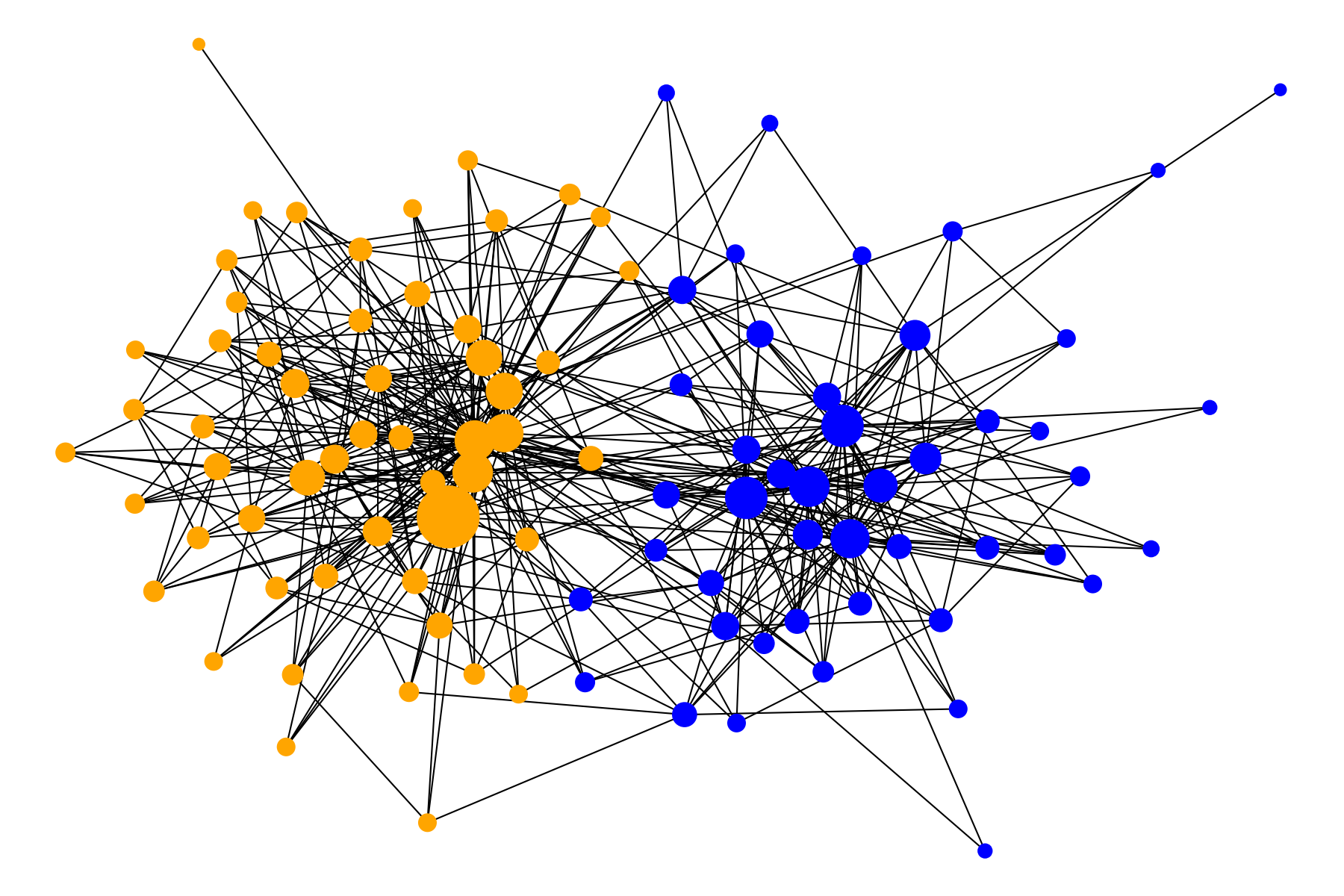}
    }\hspace{1mm}
    \subfigure[Dense and Uniform]{
        \label{fig:d_u}
        \includegraphics[width=0.22\textwidth]{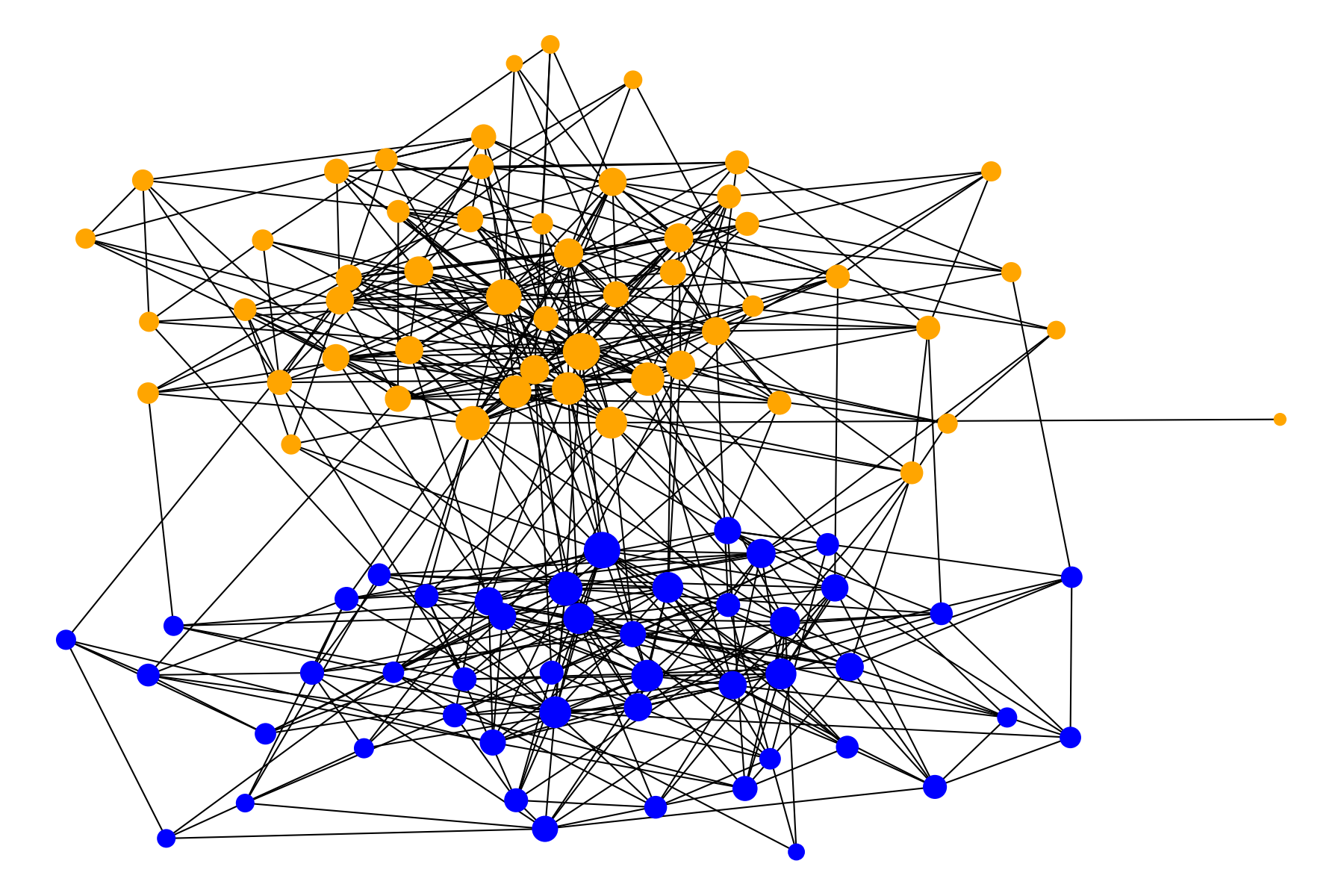}
    }\hspace{1mm}
    \subfigure[Sparse and Power-law]{
        \label{fig:s_s}
        \includegraphics[width=0.22\textwidth]{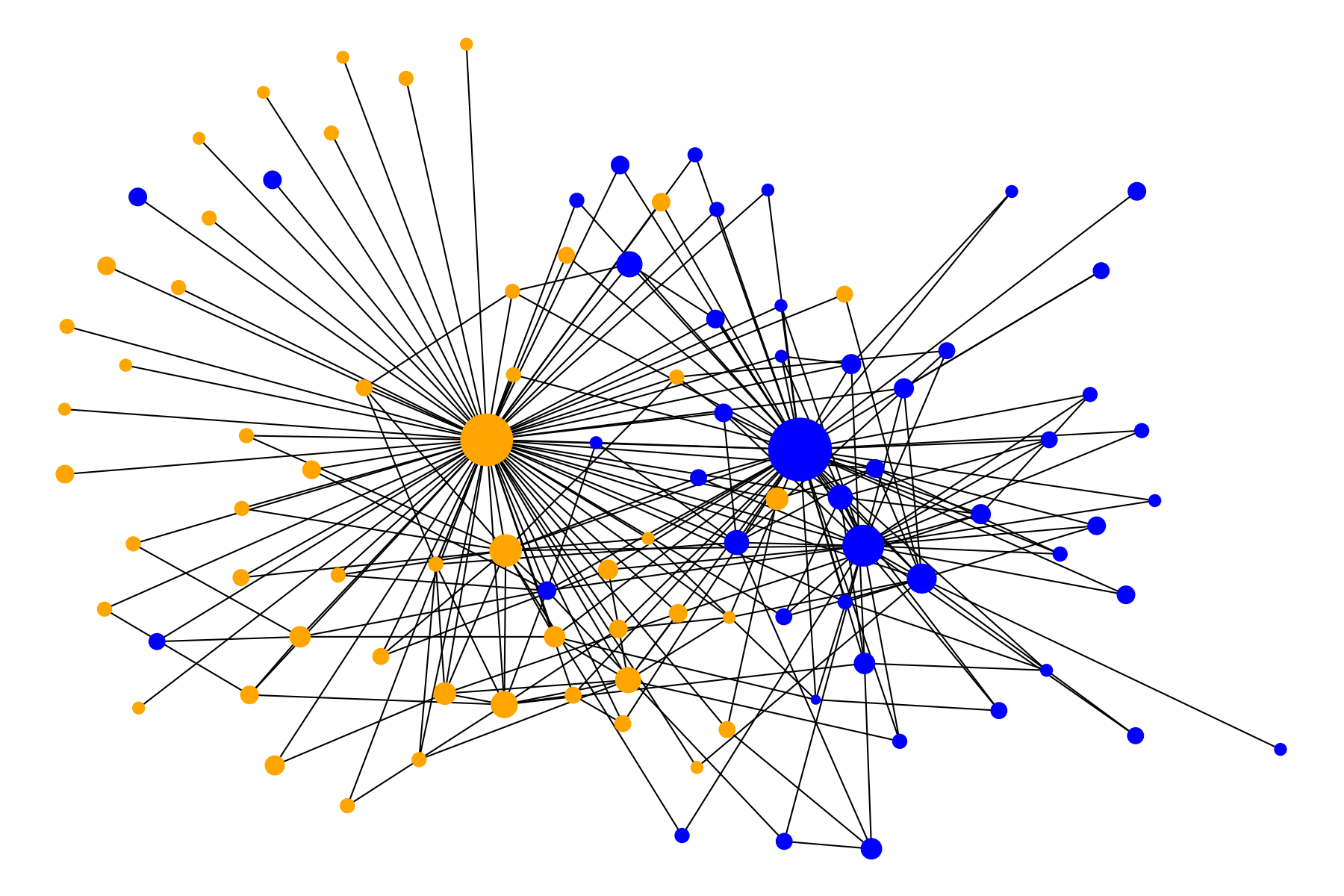}
    }\hspace{1mm}
    \subfigure[Sparse and Uniform]{
        \label{fig:s_u}
        \includegraphics[width=0.22\textwidth]{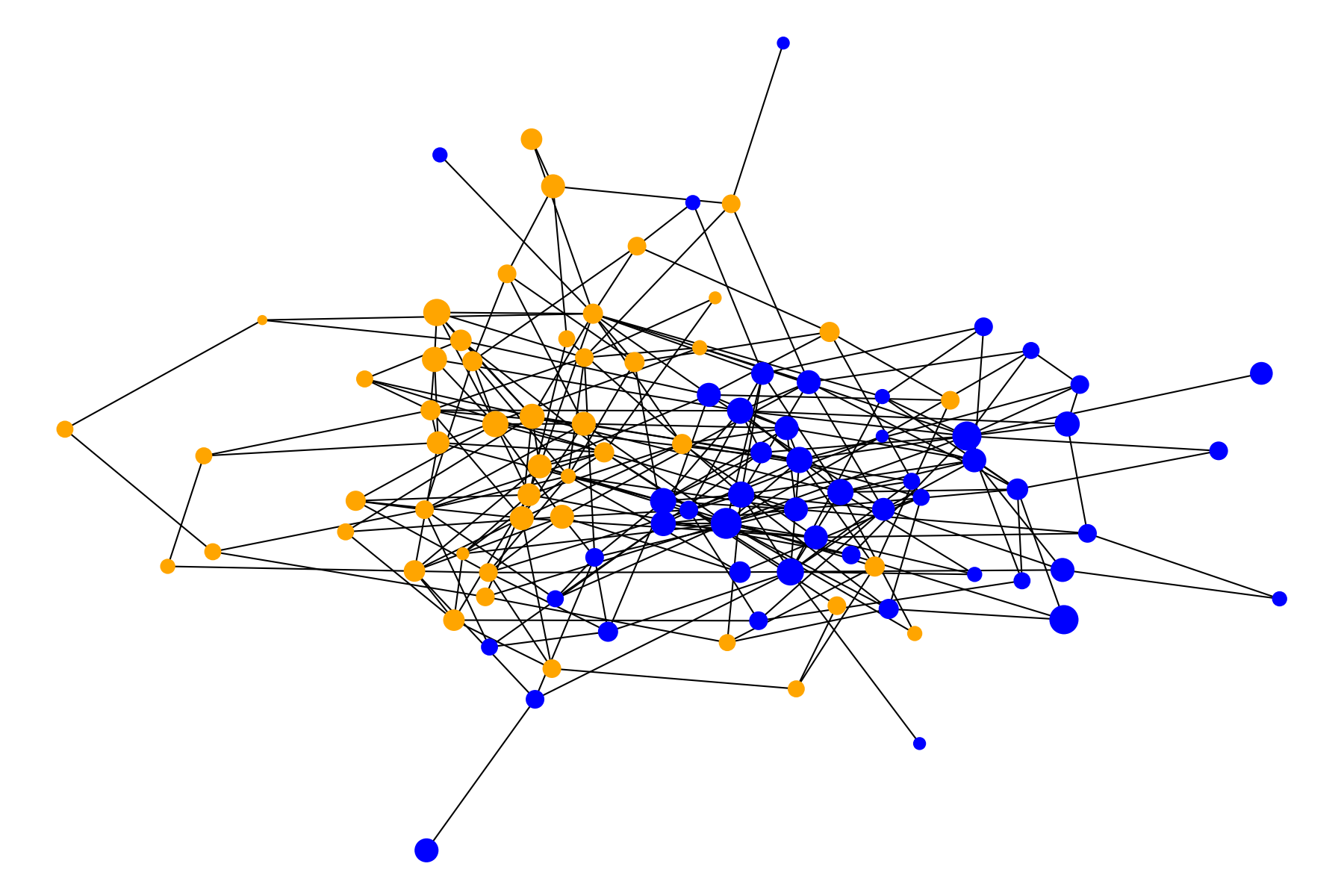}
    }
    \caption{Synthetic graphs generated by DCBM with different statistical properties}
    \label{sg}
\end{figure*}
\subsection{Pre-training procedure}
As stated in the introduction, we desire to pretrain $\cF_{\theta}$ on synthetically generated graphs, which can cover as wider range of graph property as possible. Therefore, we choose degree-corrected block model (DCBM)~\cite{holland1983stochastic}, a famous graph generation algorithm that can generate network with underlying cluster structure and controlled degree  distribution. Specifically, in DCBM, we assume there exist $K$ non-overlapping clusters $\cC$, randomly assign each node a corresponding cluster, and get an indicator function $\Ib(\cdot)$ to tell which cluster $C$ a given node $v$ belongs to. We then sample a symmetric probability matrix $P = (P_{c_1, c_2})$ denoting the probability that node in cluster $c_1$ will connect to node in cluster $c_2$, and assume nodes in the same cluster have higher probability to be connected than in different clusters. Next, DCBM has another degree-corrected probability $\theta$ to control the degree distribution of each node. Since most real-world graphs follow power-law degree distribution, we assume $\theta$ are sampled from $P(\theta) = k \theta ^{-\gamma}$. Finally, we generate the network by sampling the adjacency matrix as: $A_{uv} = A_{vu} \stackrel{iid.}{\sim} Bernoulli (\theta_u\theta_v P_{\Ib(u),\Ib(v)})$. Noted that here we have five hyper-parameters to control the generated graph, the node number $|\cV|$, total cluster number $|\cC|$, density of cluster $P$, and $k$ and $\gamma$ to control degree distribution. Thus, by enumerating all these parameters, we can generate a wide range of graphs with various properties, as is shown in Figure~\ref{sg}.
\subsection{Adaptation Procedure}
In literature~\cite{DBLP:journals/corr/abs-1903-05987}, there are two widely used methods for adapting the pre-trained model to downstream tasks: Feature-based, and Fine-Tuning. Both methods have their pros and cons. Inspired by them, in this paper we propose to combine the two methods into a more flexible way. As shown in Figure~\ref{fig:encoder}, after doing the weighted combination, we can choose a fix-tune boundary in the middle of GNNs. The GNN blocks below this boundary are fixed, while the ones above the boundary are fine-tuned. Thus, for some tasks that are closely related to our pre-trained tasks, we choose a higher boundary, and reduce the parameters to be fine-tuned, so that we can have more computation capacity left for the downstream decoder. While for tasks less related to our pre-trained tasks, we choose a lower boundary to make the pre-trained GNNs adjust to these tasks and learn task-specific representation.

%% file: section/evaluation.tex
In this section, we investigate to what extent the pre-training framework can benefit downstream tasks. We also conduct ablation study to analyze our approach.
\paragraph{Experiment Setting}
We evaluate the pre-training strategy on tasks at the levels of node, link and graph. For node classification, we utilize Cora and Pubmed datasets to classify the subject categories of a paper. For link classification, we evaluate on MovieLens 100K and 1M datasets, which aim to classify the user's rating of a movie. For graph classification, we consider IMDB-Multi and IMDB-Binary benchmarks to evaluate the model performance on classifying the genre of a movie based on the associated graph about actors and actresses.

To pre-train GNNs, we generate 900 synthetic graphs for training and 124 for validation, with node size ranges between 100-2000. The cluster assignment and centrality can be directly obtained or calculated during the graph generation process. At each step of the pre-training, we sample 32 graphs, mask $20\%$ of the links, use the masked graph as input for all the three tasks and these masked links as labels for the Denoising Link Reconstruction task. The model is optimized based on corresponding pre-training losses by the Adam optimizer. 
For downstream tasks, we model them with a standard 2-layer GCNs cascaded with the pre-trained model.
In the adaptation phase, without further mentioning, we 
fix the embedding transformation $E$ and fine-tune all the GNNs layers. For each setting, we independently conduct experiments 10 times and report mean and variance of the performance. All the results are evaluated in micro F1-score.

\begin{table}[t!]
\footnotesize
\caption{Micro F1 (\%) results on node, link and graph classification tasks. The results confirm that pre-training significantly improves the baseline without pre-training.}
\label{tab:simulate} 
\begin{tabular}{@{}c|cc|cc|cc@{}}\toprule
 \multirow{2}{*}{{Method}}  & \multicolumn{2}{c|}{Node Classification} & \multicolumn{2}{c|}{Link Classification} & \multicolumn{2}{c}{Graph Classification} \\ \cmidrule{2-7} 
             & Cora   & Pubmed   & ML-100K   & ML-1M  & IMDB-M      & IMDB-B      \\ 
            \midrule \multicolumn{7}{c}{Structure Feature Only (w/o. additional node attributes)}\\ \bottomrule
{Baseline (No Pretrain)}    & $61.9\pm 1.7$ & $61.4\pm 1.1$  &  $61.7 \pm 1.2$    &  $72.5 \pm 1.2$  &  $50.7 \pm 1.4$  &    $71.6 \pm 1.7$ \\
{Pre-train (All Tasks)}    & $\pmb{67.5\pm1.1}$ & $\pmb{66.0\pm1.4}$ &  $\pmb{71.3 \pm 1.4}$   & $\pmb{76.9 \pm 0.8}$ &  $\pmb{53.7 \pm 0.7}$  &  $\pmb{73.2 \pm 0.8}$ \\ \bottomrule
\multicolumn{7}{c}{With additional node attributes}\\
\toprule
{Baseline (Attr. Only)} & $81.3\pm 0.6$ & $78.7\pm0.7$  &  $73.3 \pm 1.4$  & $76.9 \pm 0.6$  &     /        &   /          \\ 
{Baseline (No Pretrain + Attr.)} & $78.7\pm 1.1$ & $76.5\pm0.9$  &  $72.1 \pm 1.1$  & $76.3 \pm 0.8$  &     /        &   /          \\ 
{Pre-train (All Tasks + Attr.)}    & $\pmb{81.9\pm1.0}$ & $\pmb{79.1\pm0.9}$  &  $\pmb{74.0 \pm 0.9}$  & $\pmb{77.4 \pm 0.8}$ &           /  &  /           \\
 \bottomrule
\end{tabular}
\end{table}

\begin{table}[t]
\footnotesize
\caption{Ablation study: Performance when pre-train on single task. Different pre-training tasks are beneficial to different downstream tasks according to their task properties} 
\label{tab:ablation} 
\begin{tabular}{@{}c|cc|cc|cc@{}}\toprule
 \multirow{2}{*}{{Method}}  & \multicolumn{2}{c|}{Node Classification} & \multicolumn{2}{c|}{Link Classification} & \multicolumn{2}{c}{Graph Classification} \\ \cmidrule{2-7} 
             & Cora   & Pubmed   & ML-100K   & ML-1M  & IMDB-M      & IMDB-B      \\ 
            \midrule \multicolumn{7}{c}{Structure Only}\\ \bottomrule
{Pre-train (Reconstruct)}     & $66.9\pm0.9$ & $63.8\pm1.2$  & $\pmb{72.4 \pm 1.1}$   & $\pmb{77.3 \pm 1.4}$ &   $53.9 \pm 1.2$    & $\pmb{74.1 \pm 1.4}$    \\ 
{Pre-train (Rank)}   & $66.1\pm0.7$ & $63.2\pm1.0$  &  $68.5 \pm 1.3$  & $73.6 \pm 1.2$ &          $\pmb{54.3 \pm 1.4}$   & $73.6 \pm 1.4$    \\ 
{Pre-train (Cluster)}      & $\pmb{67.9\pm0.8}$ & $65.6\pm0.9$ & $69.2 \pm 1.6$  &  $74.8 \pm 1.1$ &  $52.8 \pm 1.2$  &  $72.1 \pm 1.3$  \\ \midrule
{Pre-train (All Tasks)}    & $67.5\pm1.1$ & $\pmb{66.0\pm1.4}$ &  $71.3 \pm 1.4$   & $76.9 \pm 0.8$ &  $53.7 \pm 0.7$  &  $73.2 \pm 0.8$ \\ \bottomrule
\end{tabular}
\end{table}

\paragraph{To what extent can pre-training benefit downstream tasks?}
The first experiment attempts to verify
whether pre-training can capture generic structural features and improve performance of downstream models. To this end, we compare our approach with a baseline model whose parameters are initialized randomly. As is shown in the first block of Table~\ref{tab:simulate}, pre-training outperforms the baseline on all the six downstream tasks by 7.7\% micro-F1 in average.

Next, we show that the pre-training can as well improve state-of-the-art models when these models leverage additional node attributes as strong features. 
To this end, for the pretraining approach, we concatenate the node 
attributes with the model output, and adapt the entire model on the downstream tasks. We compare the model with an existing baseline that only takes the node attribute as the input features, and a baseline that concatenate the node attributes with a randomly initialized model without pretraining.
As is shown in the second block of Table~\ref{tab:simulate}, both the baseline models and the pre-trained model benefit from the additional node attributes, compared to the first block. And our pre-trained model consistently improve the performance even with these strong node features given.

\paragraph{Ablation Studies}
We further conduct ablation experiments to analyze the proposed approach. 


First, we analyze which \emph{pre-training task} can benefit more for a given downstream task. We follow the same setting to pre-train the GNN models on each individual task and adopt them to downstram tasks.
The results are shown in Table~\ref{tab:ablation}. As we expected, different pre-training tasks are beneficial to different downstream tasks according to their task properties. 
For example, node classification task benefits the most from the cluster preserving pre-training task, indicating that cluster information is useful to detect node's label. While link classification benefits more from the denoising link reconstruction task, as they both rely on robust representation of node pairs. Graph classification gains more from both tasks of centrality score ranking and denoising link reconstruction, as they can somehow capture the most significant local patterns. Overall, the results using all three tasks consistently improve the downstream models. 

\begin{figure}[!t]
\minipage{0.35\textwidth}
  \includegraphics[width=\linewidth]{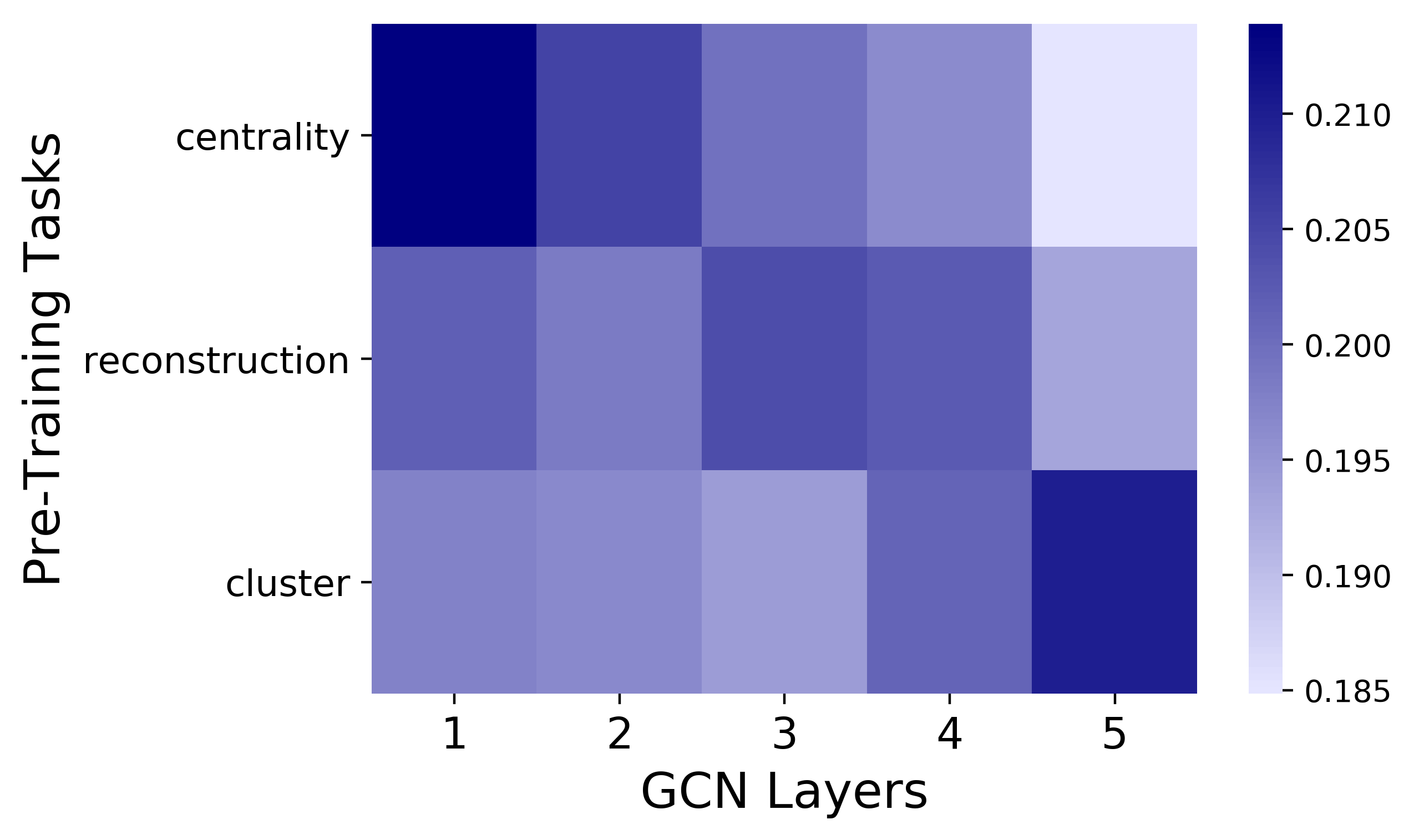}
  \caption{Attention weights $\beta^{task}_l$ of different tasks on different GNN layers through pre-training. Different tasks focus on different levels of graph structure.}\label{fig:pettern}
\endminipage\hspace{2mm}
\minipage{0.3\textwidth}
  \includegraphics[width=\linewidth]{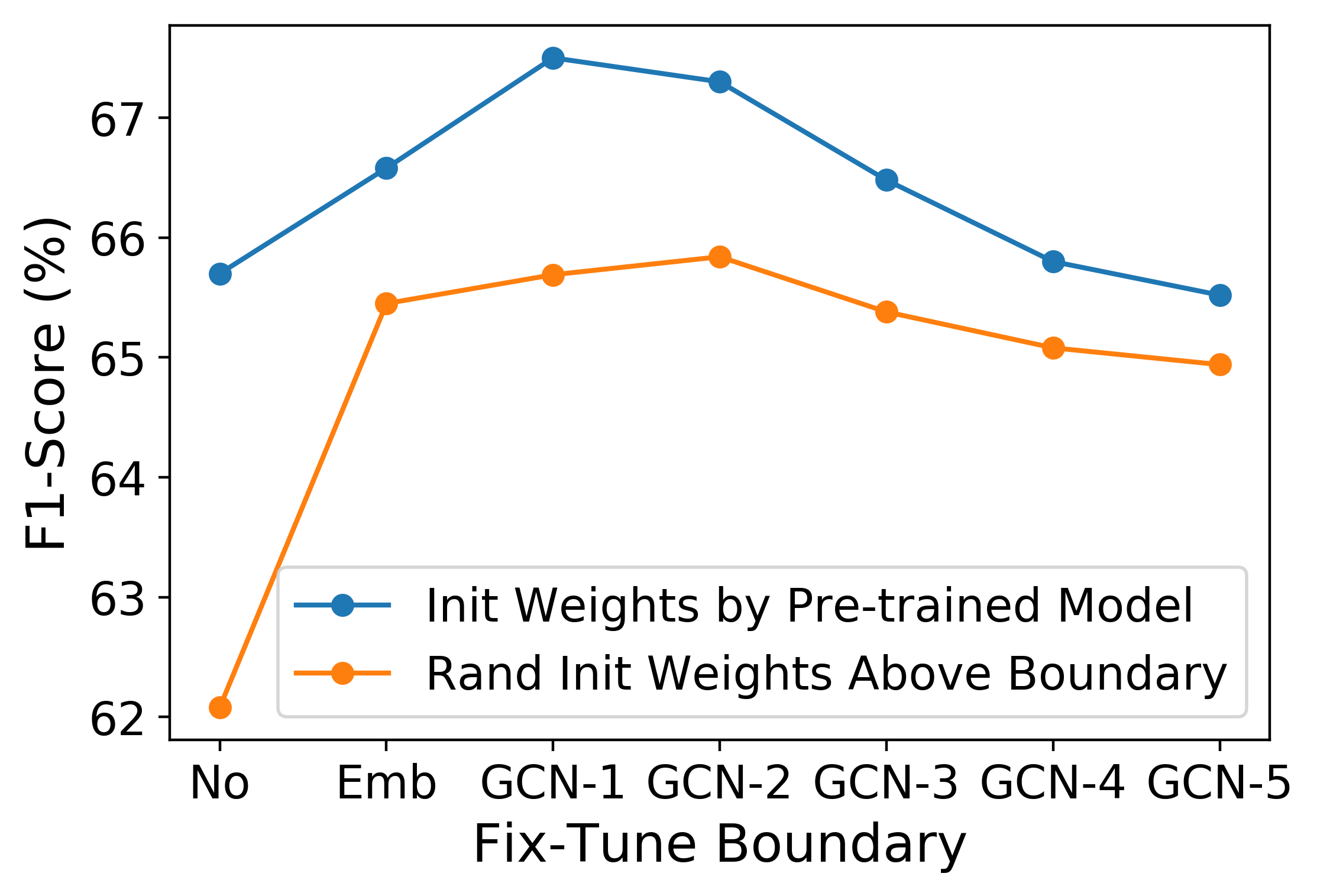}
  \caption{F1-score on Cora with different fix-tune boundary. Results show that fix the embedding and the first-layer of GNN is the best.}\label{fig:bound}
\endminipage\hspace{2mm}
\minipage{0.3\textwidth}
  \includegraphics[width=\linewidth]{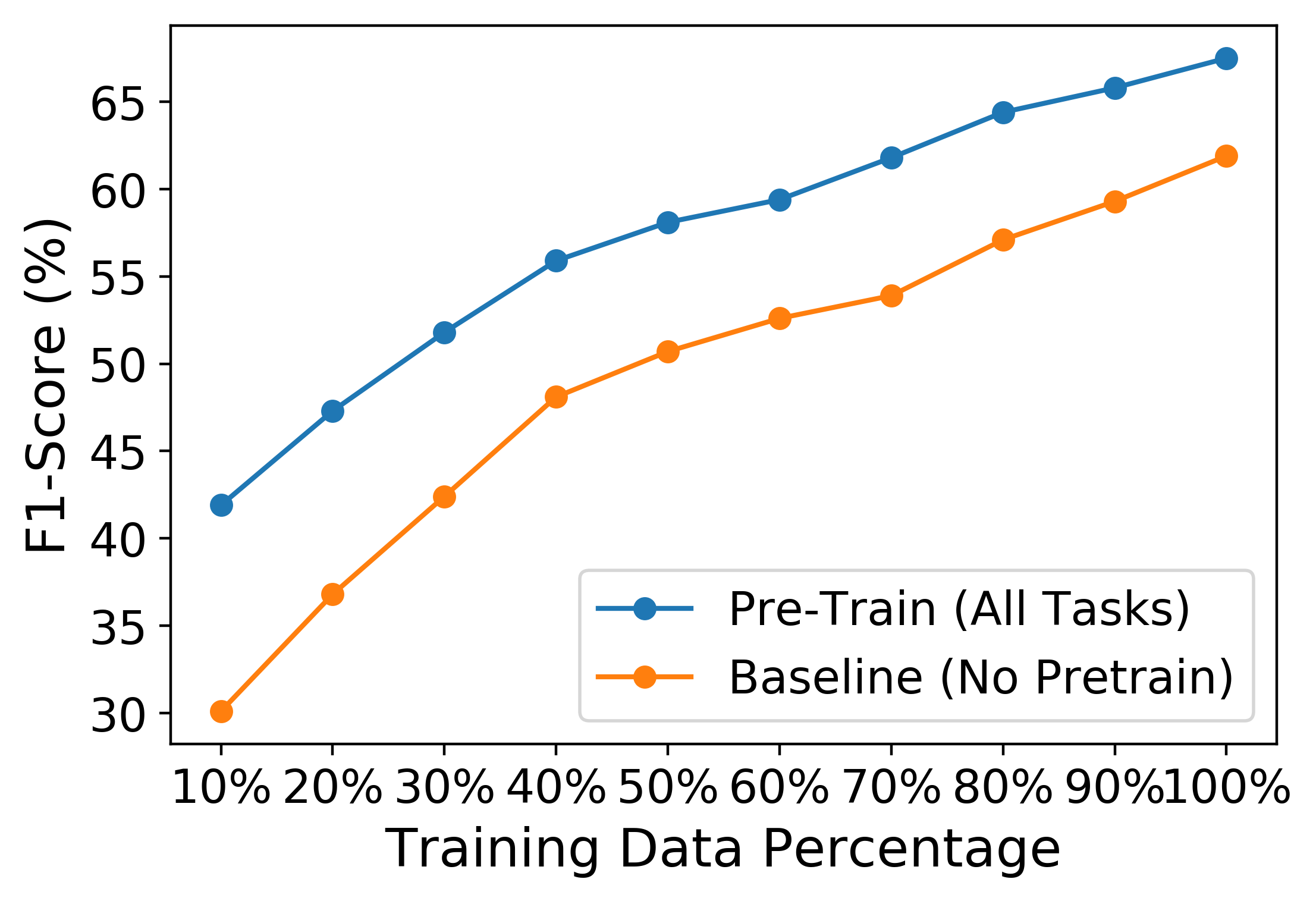}
  \caption{Benefits of pre-training on Cora with different size of training data. Pre-train improves the baseline consistently.}\label{fig:percentage}
\endminipage
\end{figure}

The fact that different pre-training tasks provide different levels of information about graphs can also be revealed by visualizing the attention weights on each GNN layer $\beta^{task}_l$. Figure~\ref{fig:pettern} shows that the cluster preserving pre-training focus mostly on high-level structural information, while denoising link reconstruction on mid-level and centrality score ranking on low-level.

Second, we investigate the \emph{adaptation strategy}. Existing techniques for adaptation ranging from fixing all the layers to fine-tuning all the layers of the pre-trained model, and it is unclear what is the best strategy in the graph setting. 
To analyze this, we treat the layers to fine-tune as a hyper-parameter and conduct experiments on the Cora dataset. Specifically, given a fix-tune boundary, we fix all the bottom layers in GNNs with fixed parameters and fine-tune the rest. 
As a reference, we also show a competitive model, where the parameters in the upper layers above the boundary are randomly initialized (c.f., initialized by the pre-trained model). 
As is shown in Figure~\ref{fig:bound}, fixing $E$ and the bottom GNN layer achieves the best performance. Meanwhile, the performance of GNNs initialized by pre-training weights outperform that of randomly initialization in all the cases. This result confirms the benefits of the pre-training framework is brought by both the fixed feature extractor and parameter initialization.

Finally, we analyze how much benefit our pre-training framework can provide given different \emph{amounts of downstream data}. 
In Figure~\ref{fig:percentage}, we show the performance difference between our approach against baseline with different amounts of training data (representing by the percentage of the training data used) on the Cora dataset. 
Results show that
the benefits of pre-training are consistent. When the training data is extremely scarce (e.g., 10\%), the improvement is more substantial.

%% file: section/conclusion.tex
In this paper, we propose to pre-train GNNs to learn generic graph structural features by using three self-supervised tasks: denoising link reconstruction, centrality score ranking, and cluster preserving. The GNNs can be pre-trained purely on synthetic graphs covering a wide range of graph properties, and then be adapted and benefit various downstream tasks. 
Experiments demonstrate the effectiveness of the proposed pre-training framework to various tasks at the level of node, link and graph. 

%% file: section/appendix.tex
\section{Implementation Details about Pre-training Framework}
In this section, we provide some implementation details of each component of our proposed pre-training framework, for ease of re-productivity.

\subsection{Details about Four Local Node Feature}
As discussed before, the input to the model should be generic to all graphs and computationally efficient to be calculated. Therefore, we choose four node-level graph properties. We list the description of them and the time complexity to calculate them for all the nodes in the graph as follow:
\begin{compactitem}
\item [i] \textbf{Degree} of a node $v$ is the number of edges linked to $v$. It is the most basic node structure property in graph theory, which can represent a node's local importance. It is also a frequently used feature for existing GNNs when there is no node attributes. After estimating the Degree, The time complexity to calculate Degree for all nodes is $\mathcal{O}(|\cV|)$. 
\item [ii] \textbf{Core Number}~\cite{DBLP:journals/corr/cs-DS-0310049} defines the connectivity of a node's subgroup. Core number of a node $v$ is the largest $K$ such that a $K$-core contains $v$, where $K$-core is a maximal subgraph such that its nodes' minimal degree is $K$. Mathematically, a subgraph $\tilde{\cG}=(\cU,\cE|\ \cU)$ induced by a subset of nodes $\cU$ is a k-core iff: $\forall u \in \cU: Deg_{\tilde{\cG}}(u) \geq k$. After estimating the Degree, the time complexity to find all k-core and assign core number to each node is $\mathcal{O}(|\cE|+|\cV|)$.
\item [iii] \textbf{Collective Influence}~\cite{DBLP:journals/corr/MoroneM15} ($CI$) defines the node's neighborhood's importance. $CI_L$ of a node $v$ in $L$-hop neighborhood is defined as: $CI_L(v) = \left(Deg(v)-1\right)\sum_{u \in N(v, L)}\left(Deg(u)-1\right)$, where $N(v, L)$ defines the $L$-hop neighborhood of $v$. In our experiment, we only calculate  $1$-hop Collective Influence ($CI_1$ where $L=1$) for computation efficiency. After estimating the Degree, the time complexity of computing $1$-hop Collective Influence for each node is $\mathcal{O}(|\cE|)$.
\item [iv] \textbf{Local Clustering Coefficient}~\cite{collect} ($CC$) of a node $v$ is the fraction of closed triplets among all the triplets containing $v$ within its $1$-hop neighborhood. $CC(v) = \frac{2 \cdot Triangle(v)}{Deg(v)\left(Deg(v)-1\right)}$. The time complexity to calculate $CC$ for each node is $\mathcal{O}(\frac{|\cE|^2}{|\cV|})$. 
\end{compactitem}
Combing all these four features together, we can quickly capture different local structural properties of each node.
Instead of Collective Influence, which we implemented on our own, the other three features are calculated using networkx package~\cite{SciPyProceedings_11}. After calculating them, we concatenate them as the 4-dim features for each node, afterward by a linear transformation $\Eb^{4\times512}$ and a non-linear transformation with $\tanh$ to get the local embedding of each node.
\subsection{Details about GNN Model Architecture}
As we need to combine the outputs of each GCN block's output, dimensions of these output vectors have to be unchanged. Therefore, both $W_1^{l}$ and $W_2^{l}$ are $512\times512$ weight matrices, so that the dimension of hidden vector $H^{l}$ at each layer is always 512.

Based on the observation reported in~\cite{DBLP:conf/aaai/LiHW18}, after applying multiple layers of GCN block, the node representations get closer. Therefore, the variance of the output nodes reduce. This observation is accordant to that the high-order Laplacian Matrix converges to a stationary distribution, so that the output of a multi-layer GCN is nearly the same. To avoid this problem, we add batch-normalization on all the nodes' embedding within one batch, so that we can concentrate on the "difference" part of each node instead of the "average" part of the whole graph. As the full-batch version of GCN training will load all the nodes in the graph, it actually does normalization among all the nodes. Suppose $x_u = \sigma \left(\Hb^{(l)}\Wb_1^{(l)}\right)\Wb_2^{(l)}[u]$, the operation of normalization is as follow:
\begin{equation}\label{def_eigv}
    \begin{aligned}
        norm(x_u) = \gamma \cdot \frac{x_u - \mu_{\cG}}{\sqrt{\sigma_{\cG}^2 + \epsilon}} + \kappa \ \ \ \ \mbox{where} \ \ \ \ \mu_{\cG} = \frac{1}{|\cV|} \sum_{u=1}^{|\cV|} x_u \ and \ \sigma_{\cG}^2 = \frac{1}{|\cV|} \sum_{u=1}^{|\cV|} (x_u-\mu_{\cG})^2
    \end{aligned}
\end{equation}

\subsection{Details about Four Node Centrality Definitions}

\begin{itemize}
    \item \textbf{Eigenvector centrality} of a node $v$ is calculated based on the centrality of its neighbors. The eigenvector centrality for node $w$ is the $w$-th element of the vector $x$ defined by the equation
        \begin{equation}\label{def_eigv}
            \begin{aligned}
                Ax = \lambda x,
            \end{aligned}
        \end{equation}
    where $A$ is the adjacency matrix of the graph with eigenvalue $\lambda$. By virtue of the Perron–Frobenius theorem, there is a unique solution $x$, all of whose entries are positive, if $\lambda$ is the largest eigenvalue of the adjacency matrix $A$~\cite{newman2010networks}. The time complexity of eigenvalue centrality is $\cO(|\cV|^3)$.
    
    \item  \textbf{Betweenness Centrality} of a node $v$ is defined as:
        \begin{equation}\label{def_betw}
            \begin{aligned}
                C_b(v) = \frac{1}{|\cV|(|\cV|-1)}\sum_{u \ne w \ne  v}\frac{\sigma_{uw}(v)}{\sigma_{uw}},
            \end{aligned}
        \end{equation}
    where $|\cV|$ denotes the number of nodes in the network, $\sigma_{uw}$ denotes the number of shortest paths from $u$ to $v$, and $\sigma_{uw}(v)$ denotes the number of shortest paths from $u$ to $w$ that pass through $v$. The time complexity of betweenness centrality is $\cO(|\cV|\cdot |\cE|)$.
    
    \item  \textbf{Closeness Centrality} of a node $v$ is defined as:
        \begin{equation}\label{def_close}
            \begin{aligned}
                C_c(v) = \frac{1}{\sum_{u}d(u,v)},
            \end{aligned}
        \end{equation}
    where $d(u,v)$ is the distance between nodes u and v. The time complexity of closeness centrality is $\cO(|\cV|\cdot |\cE|)$.
    
    \item \textbf{Subgraph centrality}: Subgraph centrality of the node $w$ is the sum of weighted closed walks of all lengths starting and ending at node $w$~\cite{estrada2005subgraph}. The weights decrease with path length. Each closed walk is associated with a connected subgraph. It is defined as:
        \begin{equation}\label{def_subgraph}
            \begin{aligned}
                C_{sc}(w) = \sum_{j=1}^{N}(v_j^w)^2e^{\lambda_j},
            \end{aligned}
        \end{equation}
        where $v_j^w$ is the $w$-th element of eigenvector $v_j$ of the adjacency matrix $A$ corresponding to the eigenvalue $\lambda_j$. The time complexity of subgraph centrality is $\cO(|\cV|^4)$.
\end{itemize}

\subsection{Details about Decoders for the three tasks}
Both $\cD^{rec}$ and $\cD^{cluster}$ are implemented by pairwise Neural Tensor Network (NTN)~\cite{DBLP:conf/nips/SocherCMN13}. Mathematically:
\begin{equation}
         NTN(x_i, x_j)=\tanh\left(x_i^T W^{[1:4]} x_j+V\left[
        \begin{matrix}
        x_i \\
        x_j,
        \end{matrix}
        \right]+b\right),
\end{equation}
where $x_i$ and $x_j$ are two representation vector for the pair input with dimension $512$ (in our case, it's node embedding of GNN output or cluster embedding which aggregates multiple node embeddings) and the output of $NTN$ is a $4$-dim vector. $x_i^T W^{[1:4]} x_j$ is a bilinear tensor product, results in a $4$-dim vector, which model the interaction of $x_i$ and $x_j$. Each entry of this vector is computed by one slice of the tensor $W^{[1:4]} \in \mathbb{R}^{512\time512\time4}$, resulted in $x_i^T W^{[i]} x_j$, where $i \in [1,4]$. The second part is a linear transformation of the concatenation of $x_i$ and $x_j$, with $V \in \mathbb{R}^{4\times1024}$, adding a bias term $b\in\mathbb{R}^4$.

After getting the $4$-dim vector output, we can cascade it by an output layer to conduct classification for Denoising Link Reconstruction and Cluster Preserving.

For Centrality Score Ranking, we simply decode the node embedding into a two-layer MLP ($512-256-1$), so as to get one digit output as the rank score of each centrality metric. We thus can cascade it with a pair-wise ranking layer to conduct the task.

\subsection{Details about Pre-training Procedure}

As is discussed above, we choose degree-corrected block model (DCBM)~\cite{holland1983stochastic} to generate network with underlying cluster structure and controlled degree  distribution. Specifically, we generate the graph with adjacency matrix as $A_{uv} = A_{vu} \stackrel{iid.}{\sim} Bernoulli (\theta_u\theta_v P_{\Ib(u),\Ib(v)})$. 

There are two parts we can control the generation process, the symmetric probability matrix of cluster connectivity: $P = (P_{c_1, c_2})$ and degree-corrected term $\theta$ following $P(\theta) = k \theta ^{-\gamma}$. For cluster matrix, we sample the total cluster number $|\cC|$ randomly from 2 to 10, with a parameter $p\_div\_q$ to control how much the ratio is the in-cluster probability divided by the inter-cluster probability, which is sampled from 3 to 6. For degree-corrected term $\theta$, we sample $k$ from 0.1 to 2, which controls the connectivity of the graph, where a higher $k$ makes the graph denser, and vice versa. We then sample $\gamma$ from 2 to 10, which controls how much the degree distribution fits the power-law distribution, where a higher $\gamma$ will make the degree distribution more uniform, and vice versa. Utilizing all these parameters, we generate 1024 graphs with different graph structure properties, where the first 900 for training and the remaining 124 for validation. After that, the details of the pre-training procedure is described in Algorithm~\ref{alg:Framwork}, where we iteratively sample part of the graphs, mask some links out and train the model using three self-supervised loss.

\begin{algorithm}[tb] 
\caption{Pre-Training Procedure} 
\label{alg:Framwork} 
\begin{algorithmic}[1] 
\REQUIRE
Synthetically Generated Graphs Dataset $\cD = \{\cG\}$
\STATE Training Graph Dataset $TD = \cD[:900]$, Validation Graph Dataset $VD = \cD[900:]$
\WHILE {not converge}
    \STATE Sample a subset of graphs $S={\cG^*}$ from Training Graph Dataset $TD$ and remove $20\%$ of the link from each graph.
    \FOR {$\cG^* \in S$}
        \STATE Sample 128 node pairs that are previously connected but masked out, and 256 node pairs that are not connected, as the training samples. 
        \STATE Use the node pairs to calculate $\cL_{rec}$ and  $\cL_{rank}$. Use the nodes within these pairs and half of there contexts as cluster support to calculate $\cL_{cluster}$. 
        \STATE Sum the three loss to get $\cL$, and update the model parameters using ADAM optimizer with a learning rate of 0.001.
    \ENDFOR
    \STATE Evaluate the model using the three Task, pick the best model on Validation Graph Dataset $VD$.
\ENDWHILE
\RETURN Best model with highest validation performance.
\end{algorithmic} 
\end{algorithm}

\section{Details about Experimental Datasets}
The Datasets statistics of our three different type of downstream tasks are summarized as Table \ref{data_stats}.

\begin{table}[H]
\caption{Experimental Dataset Statistics}\label{data_stats}
\resizebox{\linewidth}{!}{%
\begin{tabular}{@{}lllrrccccc@{}}
\toprule
\textbf{Task} &\textbf{Data} &\textbf{Type} & $\#$\textbf{Graphs} & $\#$\textbf{Nodes} &$\#$\textbf{Edges} &$\#$\textbf{Classes}  & \textbf{Node Attributes} &\textbf{Train/Test rate} \\ \midrule

\multirow{2}{*} {\begin{tabular}[c]{@{}l@{}}Node \\ Classification\end{tabular}}  
        & Cora & Citation & 1& 2,708 & 5,429 & 7  & bag-of-words & 0.052 \\
        & Pubmed & Citation & 1& 19,717 & 44,338 & 3 & bag-of-words & 0.003 \\ \midrule
            
\multirow{2}{*} {\begin{tabular}[c]{@{}l@{}}Link \\ Classification\end{tabular}}  
        & ML-100K & Bipartite & 1& 2,625 & 100,000 & 5 &  User/Item Tags & 0.8  \\
        & ML-1M & Bipartite & 1& 9,940 & 1,000,209 & 5 &  User/Item Tags  & 0.8\\ \midrule
            
\multirow{2}{*} {\begin{tabular}[c]{@{}l@{}}Graph \\  Classification\end{tabular}} 
        & IMDB-M  & Ego & 1500 &  13.00& 65.94 &  3  & / & 0.1   \\
        & IMDB-B  & Ego & 1000 &  19.77& 96.53 &  2  & / & 0.1   \\ \bottomrule
\end{tabular}}
\end{table}

\paragraph{Citation networks} We consider two citation networks for the node classification tasks: Cora and Pubmed. For these networks, nodes represent documents, and edges denote citations between documents. Each node contains a sparse bag-of-words feature vector representing the documents. 

\paragraph{MovieLens(ML) networks} We consider two movielens networks for the link classification task: ML-100K and ML-1M, where each node represents a user or a movie, and each edge denotes a rating of a movie made by a user. ML-100K was collected through the MovieLens web site (movielens.umn.edu) during the seven-month period from September 19th, 1997 through April 22nd, 1998. This data has been cleaned up - users who had less than 20 ratings or did not have complete demographic information were removed from this data set. ML-100K contains 100,000 ratings (1-5) from 943 users on 1682 movies. ML-1M contains 1,000,209 anonymous ratings of approximately 3,900 movies made by 6,040 MovieLens users who joined MovieLens in 2000.

\paragraph{IMDB networks} We consider using IMDB networks, which is a widely used graph classification dataset, for evaluation. Each graph in this dataset corresponds to an ego-network for each actor/actress, where nodes correspond to actors/actresses and an edge is drawn between two actors/actresses if they appear in the same movie. Each graph is derived from a pre-specified genre of movies, and the task is to classify the genre that graph it is derived from. This dataset doesn't have explicit features, so normally people only use the degree as an input feature. IMDB-M denotes for the multi-class classification dataset, and IMDB-B denotes for the binary classification dataset.